\documentclass{article}

\usepackage{arxiv}

\usepackage[utf8]{inputenc} 
\usepackage[T1]{fontenc}    
\usepackage{hyperref}       
\usepackage{url}            
\usepackage{booktabs}       
\usepackage{amsfonts}       
\usepackage{nicefrac}       
\usepackage{microtype}      
\usepackage{lipsum}		
\usepackage{graphicx}
\usepackage{natbib}
\usepackage{doi}

\usepackage{wrapfig}
\usepackage{caption}
\usepackage{subcaption}
\usepackage{amsmath,amssymb,amsfonts}
\usepackage{algorithmic}
\usepackage{graphicx}
\usepackage{textcomp}
\usepackage{xcolor}
\usepackage{tabto}
\usepackage{comment}
\usepackage{booktabs}
\usepackage{diagbox}
\usepackage{flexisym}
\usepackage{bbding}
\usepackage{multirow}
\usepackage{makecell}
\usepackage{amsmath, bm}
\usepackage{tikz}
\usepackage{eso-pic}
\usepackage{algorithm}
\usepackage{algorithmic}
\usepackage{wrapfig}
\usepackage{bbding}
\usepackage{stfloats}
\usepackage{colortbl}
\usepackage{threeparttable}
\usepackage{ulem}

\title{WikiStyle+: A Multimodal Approach to Content-Style Representation Disentanglement for Artistic Image Stylization}

\author{  
    Zhuoqi Ma ,
    Yixuan Zhang , 
    Zejun You ,
    Long Tian ,
    Xiyang Liu \\
    School of Computer Science and Technology\\
    Xidian University\\
    Xi'an, Shaanxi, 710126, China \\
    zhuoqima@xidian.edu.cn
}  

\date{}

\renewcommand{\headeright}{Technical Report}
\renewcommand{\undertitle}{Technical Report}
\renewcommand{\shorttitle}{\textit{WikiStyle+}}

\begin{document}
\maketitle

\begin{abstract}
Artistic image stylization aims to render the content provided by text or image with the target style, where content and style decoupling is the key to achieve satisfactory results. However, current methods for content and style disentanglement primarily rely on image supervision, which leads to two problems: 1) models can only support one modality for style or content input;2) incomplete disentanglement resulting in content leakage from the reference image. To address the above issues, this paper proposes a multimodal approach to content-style disentanglement for artistic image stylization. We construct a \textit{WikiStyle+} dataset consists of artworks with corresponding textual descriptions for style and content. Based on the multimodal dataset, we propose a disentangled representations-guided diffusion model. The disentangled representations are first learned by Q-Formers and then injected into a pre-trained diffusion model using learnable multi-step cross-attention layers. Experimental results show that our method achieves a thorough disentanglement of content and style in reference images under multimodal supervision, thereby enabling more refined stylization that aligns with the artistic characteristics of the reference style. The code of our method will be available upon acceptance.
\end{abstract}

\section{Introduction}
Artistic image stylization task aims at creating new images by applying artistic styles to content. In stylization, the concept of ``content'' is well-defined, typically referring to the subject/semantics of the input image or text. However, the definition of ``style'' is relatively vague and lacks consistent standards. Art historian, Meyer Schapiro, has defined artistic style as: ``The constant form and sometimes the constant elements, qualities, and expression in the art of an individual or a group'' ~\citep{karkov2003anglo}. For example, Impressionism emphasizes the natural representation of light and color. Therefore, to truly capture the artistic characteristics of the reference style, stylization models needs to effectively disentangle and control content and style representation during the stylization process.

\begin{figure*}[htbp]
    \centering
    \includegraphics[width=\textwidth]{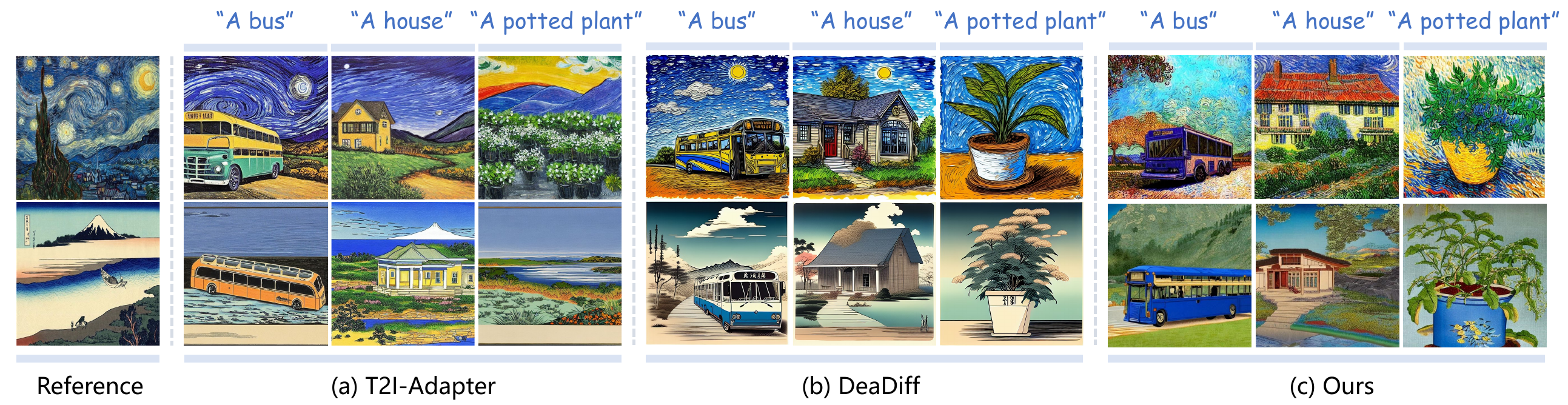}
    \caption{Given a style reference image, our model can generate artistic images with refined stylization, effectively capturing the distinctive artistic characteristics of the intended style.} 
    \label{fig:motivation}
\end{figure*}

Recently, diffusion models~~\citep{ho2020denoising,rombach2022high} have demonstrated great potential in text-to-image stylization tasks ~~\citep{ramesh2022hierarchical, ye2023ip, mou2024t2i,chen2024artadapter} with their powerful generative capabilities. These methods typically extract reliable features in the reference images serving as conditional information to guide the diffusion model to follow the predetermined style. However, the features extracted by the encoder often couple style and semantics. This, in turn, results in content leakage within the stylized output, where elements from the reference image appear in the generated result despite being inconsistent with the intended subject, as shown in Fig. \ref{fig:motivation} (a).

Some approaches ~\citep{xing2024csgo}~\citep{qi2024deadiff} attempt to achieve disentanglement by using separate encoders to extract style and content representations. However, they still face the following challenges: 1) These methods depend on AI-generated paired content and stylized images rather than artistic works, which in turn causes the generated results resemble cartoon-like visuals rather than authentic oil paintings (Fig.\ref{fig:motivation}(b)). Moreover, they rely solely on image-based supervision, limiting the model to image inputs. 2) These methods overlook the fact that artists adapt their style based on different subjects, leading to generated images that exhibit similar colors and textures regardless of the theme (Fig.\ref{fig:motivation}(b)). This inconsistency fails to reflect the way real artworks vary in style across different subjects. 

To overcome these challenges, we introduces a multimodal approach for content-style representation disentanglement. First, we constructed a multimodal artistic dataset, WikiStyle+. We curated authentic art images from the WikiArt website along with their associated style information, including artist names, genres, and painting mediums, to serve as references for style descriptions. Additionally, we utilized a large language model to generate textual descriptions of the content depicted in these artworks. In this way, we addressed the lack of explicitly disentangled style and content data in artistic images from a multimodal perspective. Based on the constructed WikiStyle+ dataset, we proposed a content-style disentangled representation-guided diffusion model. Through multimodal alignment tasks, the Q-former aligns the learned image style features with style descriptions and the learned content features with content descriptions. This method utilizes multimodal data to provide the style and content information of the reference images for disentanglement supervision, achieving explicit separation of content and style information from art images. Building on this, we inject the learned style and content representations into the multi-step cross-attention layers of the diffusion model, leveraging its generative capabilities to achieve image stylization and generation.


With the explicit disentanglement of style and content, our method adapts visual elements based on the subject and varying style prompts, moving beyond simply replicating the color palette of the reference image. This enables more nuanced stylization that better captures the artistic characteristics of the intended style. As shown in Fig. \ref{fig:motivation} (c), our method captures broader brushstroke techniques of the artist rather than simply replicating signature swirling patterns. Moreover, the multimodal disentanglement supervision enables style and content inputs from both image and text modalities, overcoming the limitations of previous methods that rely solely on image-based supervision for decoupling. In summary, our contributions are threefold:

\begin{itemize}

    \item We constructed \textbf{WikiStyle+}, a multimodal artistic dataset, to address the lack of explicitly disentangled style and content data from a multimodal perspective.
    \item We propose a multimodal approach for the explicit disentanglement of style and content representations through multimodal supervision, enabling the model to accept diverse modality inputs for artistic image stylization.
    \item We proposed a disentangled representation-guided diffusion model, where the disentangled content and style representations are injected into the cross-attention layers at different time steps of the diffusion model, enabling more refined stylization that aligns with the artistic characteristics of the reference style.
\end{itemize}

\section{Related Work}

\subsection{Diffusion-based image stylization with multimodal latents}

In recent years, Diffusion Probabilistic Models \cite{sohl2015deep} have shown great potential in image generation \cite{dhariwal2021diffusion, ho2020denoising, song2020denoising}. With the advancement of large-scale multimodal pre-trained models \cite{radford2021learning, li2023blip}, diffusion models have achieved remarkable success in text-to-image generation \cite{ramesh2022hierarchical, rombach2022high, saharia2022photorealistic}. Text-to-image stylization using diffusion models primarily falls into two categories: Optimization-based methods, which fine-tune diffusion models for specific styles \cite{ruiz2023dreambooth, kumari2023multi, sohn2023styledrop} or use textual inversion to refine text embeddings for improved style fidelity \cite{gal2022image, zhang2023inversion}.
Conditional diffusion models with pre-trained encoders, which extract stylistic features from reference images \cite{huang2023composer, li2024blip, mou2024t2i, wang2023styleadapter, ye2023ip, zhao2024uni} and inject them as conditions into the diffusion model to generate images in the desired style.

\subsection{Content-Style Disentanglement}

In image stylization, content and style disentanglement is crucial for accurately transferring the target style. Text inversion methods, such as InST ~\citep{zhang2023inversion}, VCT ~\citep{cheng2023general}, DreamBooth ~\citep{ruiz2023dreambooth}, and ArtBank ~\citep{zhang2024artbank}, map reference images to the embedding space of special text tokens via a reversal module. They construct text prompts to separately provide content and style. Unoptimized attention-based methods, like StyleAligned ~\citep{hertz2024style} and Visual Style Prompting ~\citep{jeong2024visual}, achieve zero-shot content-style separation by modifying attention mechanisms in Stable Diffusion (SD). Cross-domain alignment approaches, such as StyleDiffusion ~\citep{wang2023stylediffusion} and OSASIS ~\citep{cho2024one}, leverage cross-domain and intra-domain losses in CLIP space to decouple style and content at a latent level. Adapter-based methods, including IP-Adapter ~\citep{ye2023ip} and StyleAdapter ~\citep{wang2023styleadapter}, fine-tune specific model layers or channels to separate and blend style-content without altering the model's core structure. Similarly, encoder-based adapters, like InstantStyle ~\citep{wang2024instantstyle} and DEA-Diffusion ~\citep{qi2024deadiff}, disentangle style-content within the encoding space. Despite these advances, existing methods fail to achieve explicit disentanglement in latent space, leading to content leakage, where unintended elements from the reference image persist in the generated output.

\section{\textit{WikiStyle+} Dataset}
\label{sec:database}

A major challenge in content-style disentanglement is the lack of proper supervision. Existing methods rely on AI-generated paired content and stylized images rather than genuine artistic works. However, real artworks typically do not have corresponding content images paired with them, making supervised disentanglement more difficult. In this paper, we propose a "content description"–"style description"–"artwork" triplet dataset to address the lack of explicitly disentangled style and content data from a multimodal perspective. Formally, the construction of the paired datasets involves the following three steps:



\begin{wrapfigure}{r}{0.48\textwidth}
\centering
\vspace{-6mm}
    \centering
    \includegraphics[width=0.48\textwidth]{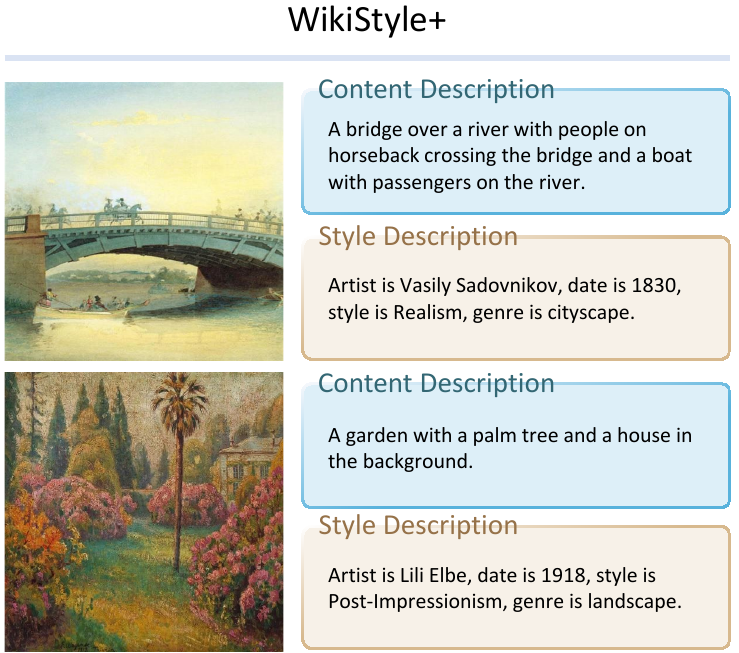}
    \caption{\small Examples from WikiStyle+ dataset, each item contains artwork, content text and style text.}
    \vspace{-5mm}
\label{tab-data}
\end{wrapfigure}

\textbf{Step 1. Image collection.} We collected 189,631 entries from WikiArt \cite{wikiart}, encompassing diverse artists and art movements. Each entry includes an artwork image and style-related metadata such as the artist, style, genre, medium, and other relevant attributes. 

\textbf{Step 2. Dataset Refinement.} To ensure stylistic clarity, we removed artworks lacking distinct stylistic features, such as those in photography, architecture, design drawings, and advertisements. Additionally, we excluded works without clear subjects, including those from abstract and minimalism movements. After this filtering process, we obtained a refined dataset of 146,547 records. For style attributes, we selected the four most prevalent and distinctive ones: artist, artistic style, genre, and medium.

\textbf{Step 3. Content description generation.} To construct paired content and style data, we utilized InternVL-Chat \cite{chen2024internvl} with the prompt, ``\textless image\textgreater, describe the content of this picture briefly." to generate content descriptions for each artwork. The generated content spans a diverse range of themes, including portraits, still lifes, natural landscapes, and cultural landscapes.  

In this way, we construct a content-style dataset based on real artistic images from a multimodal perspective, addressing the issue of insufficient explicit data for content-style disentanglement. Further details on WikiStyle+ are provided in Supplementary Material Sec.1.

\section{Method}

\begin{figure*}
    \centering
    \includegraphics[width= \textwidth]{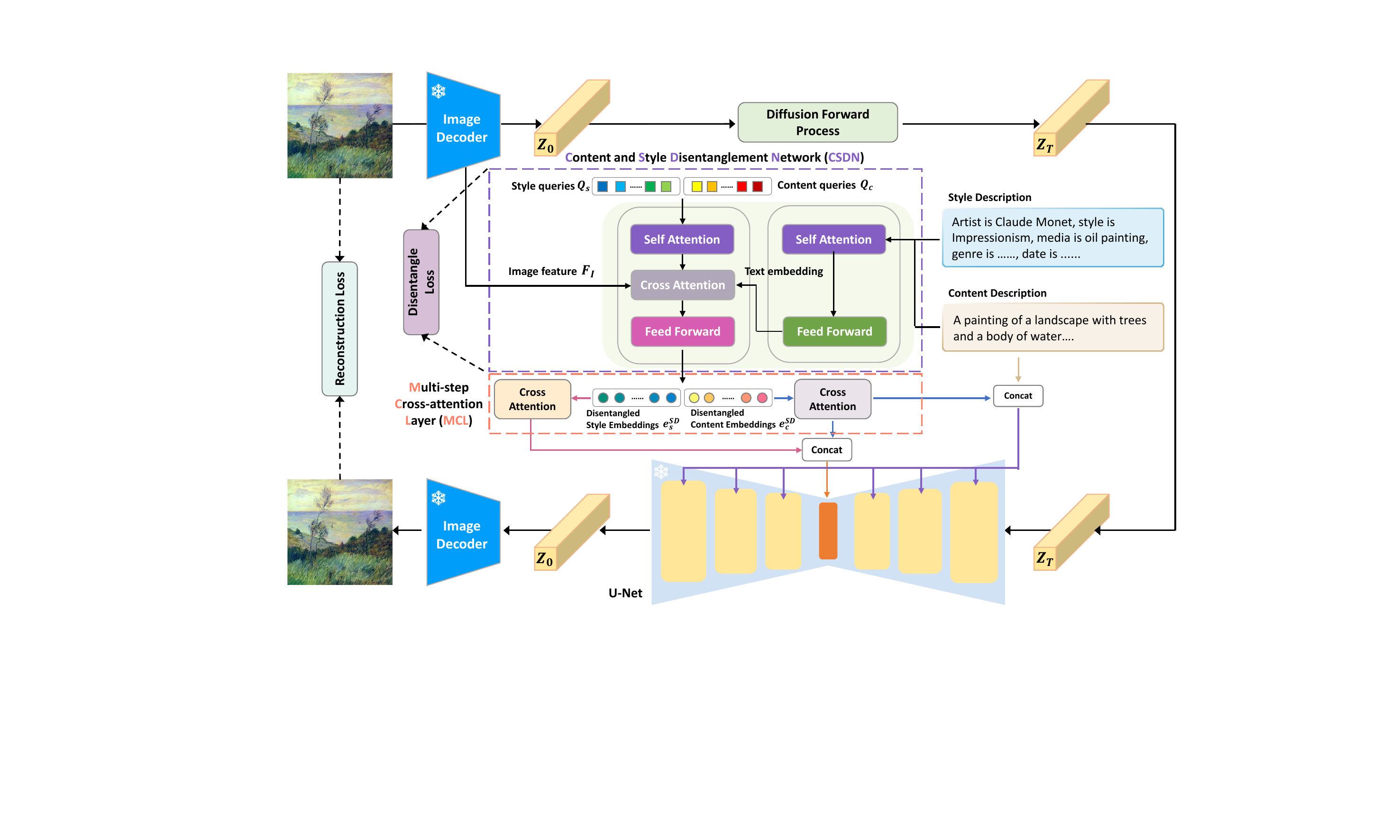}
    \caption{Overview of our model that contains three parts: 1) a pre-trained image encoder; 2) a Content and Style Disentangled Network (CSDN) with a connection to a pre-trained Stable Diffusion (SD) model; 3) a learnable multi-step cross-attention layers (MCL) to separately inject the content and style features into the SD model.}
    \label{fig:model}
\end{figure*}

We propose a disentangled content and style guided diffusion model for controllable artistic image stylization and generation, as shown in Fig. \ref{fig:model}. In Sec. \ref{sec: disentanglement}, we present our proposed Content and Style Disentangled Network (CSDN) based on a pre-trained autoencoder \cite{oord2018neuraldiscreterepresentationlearning}, it outputs disentangled content and style representations for the followup diffusion model. In Sec. \ref{sec:inject}, we introduce the Multi-step Cross-attention Layers (MCL) for controllable artistic image stylization and generation by injecting the disentangled representations into a pre-trained Stable Diffusion (SD) \cite{rombach2022high} model.

\subsection{Content and Style Disentanglement Network}
\label{sec: disentanglement}
The core of CSDN lies in the disentangled representation learning, where we employ Q-Former \cite{li2023blip} to separate the style and content from images and align the feature spaces of images and text accordingly. Different from \cite{qi2024deadiff} that uses two independent Q-Formers, we adopt a simpler design with two sets of learnable query embeddings, one dedicated to extracting content embeddings from multimodal inputs and the other for extracting style embeddings. There are two advantages by doing so: 1) Smaller model and faster convergence rate; 2) Physical decoupled query embeddings are crucial for explicit disentanglement of content and style.

The dataset is structured as triplets as introduced in Sec. \ref{sec:database}, which gives us data basis for disentangling content and style altogether. 
We achieve content and style disentanglement by minimizing an objective function that incorporates the Image-Text Contrastive Learning Loss $\mathcal{L}_{itc}$, Image-Text Matching Loss $\mathcal{L}_{itm}$, and Image-grounded Text Generation Loss $\mathcal{L}_{itg}$ as follows: 
\begin{align}
    &\mathcal{L} = \mathcal{L}_{s} + \mathcal{L}_{c}, \notag \\ 
    &\mathcal{L}_{s} = \mathcal{L}^s_{\mathit{itc}} + \mathcal{L}^s_{\mathit{itm}} + \mathcal{L}^s_{\mathit{itg}}, \quad  
    \mathcal{L}_{c} = \mathcal{L}^c_{\mathit{itc}} + \mathcal{L}^c_{\mathit{itm}} + \mathcal{L}^c_{\mathit{itg}}
\end{align}
where the superscripts $s$ and $c$ separately denote style and content, $\mathcal{L}_{s}$ is the total style loss and $\mathcal{L}_{c}$ is the total content loss. The overall objective corresponds to the disentangle loss in Fig. \ref{fig:model}.

\textbf{Image-Text Contrastive Learning.} The image is processed by a pre-trained image encoder to obtain image features $F_I$. The queries $Q$ composed of style queries and content queries then utilize the Q-Former to extract the visual representations of style $I_{s}$ and content $I_{c}$ from $F_I$, as well as the textual representations of style $T_{s}$ and content $T_{c}$ from the style and content text, respectively. Since the queries contain multiple output embeddings, we apply a pooling operation to them. Finally, we align visual representations $I_{s}$ and $I_{c}$ with textual representations $T_{s}$ and $T_{c}$ respectively.
\begin{align}
    \mathcal{L}_{\mathit{itc}} = -\frac{1}{N} \sum_{n=1}^{N} \Bigg( 
    & \log \frac{\exp\left( \mathit{d}(I_n, T_n) / \tau \right)}{\sum_{j=1}^{N} \exp\left( \mathit{d}(I_n, T_j) / \tau \right)} \notag \\ 
    & + \log \frac{\exp\left( \mathit{d}(T_n, I_n) / \tau \right)}{\sum_{j=1}^{N} \exp\left( \mathit{d}(T_n, I_j) / \tau \right)} 
    \Bigg)
\end{align}
where $\mathit{d}(\cdot,\cdot)$ denotes the cosine distance, $\tau$ is a temperature scaling parameter, $N$ is the batch size. ITC enables the model to disentangle style and content effectively by easuring that features corresponding to style and content are aligned with their respective textual descriptions.

\textbf{Image-Text Matching.} ITM operates as a binary classification task, predicting whether an image-text pair is a positive or negative match. This enables the model to focus on fine-grained correspondence between images and text. ITM computes the cosine similarity between image embedding and text embedding, then using a linear layer to map the cosine similarity into matching probability. ITM uses binary classification loss to optimize the Q-former and classifier:
\begin{align}
    \mathcal{L}_{\mathit{itm}} = -\frac{1}{N} \sum_{n=1}^{N} \Big[ 
    & y_n \log P(y_n=1 |\mathit{Pair}_{n}) \notag \\ 
    & + (1 - y_n) \log P(y_n=0 | \mathit{Pair}_{n}) 
    \Big]
\end{align}

where $\mathit{Pair}_{n}$ represents the $n$-th image-text pair, $y_n$ is the ground truth label indicating whether the i-th image-text pair is a match ($y_n = 1$) or not ($y_n = 0$), $P(y_n=1 | \mathit{Pair}_{n})$  is the model’s predicted probability that the image-text pair is a match, $N$ is the batch size.
 
\textbf{Image-grounded Text Generation} trains the model to generate coherent style and content descriptions for given input image by predicting the next word based on the extracted embeddings $I_{s}$ and $I_{c}$ from the image using queries $Q$. A lightweight text decoder is used to generation the text sequence. It consists of two main components: a transformation module that applies a dense projection, an activation function, and layer normalization to refine hidden states, and a decoder layer that maps the processed hidden states to vocabulary logits using a linear layer.
\begin{align}
P(w_m \mid w_1, w_2, \dots, w_{m-1}, I) = \text{Decoder}(h_m)
\end{align}
where $h_m = f(w_1, w_2, \dots, w_{m-1}, I)$ represents the hidden state at step
$m$, generated based on the previous words and the extracted embeddings. At each step $m$, the decoder predicts the probability distribution for the next word, given the previously generated words and the extracted embeddings I.

ITG is implemented as the cross-entropy loss between the predicted probabilities and the groundtruth sequence: 
\begin{align}
\mathcal{L}_{\mathit{itg}} = -\sum_{m=1}^{M} \log P_{\theta}(w_m | I, w_{<m})
\end{align}
where $P_{\theta}(w_m | I, w_{<m})$ is the probability of generating the next word $w_m$ given the image features $I$ and the preceding words $w_{<m}$, $M$ is the length of the text sequence. ITG encourages the model to learn robust textual representations of visual information, ensuring that the style and content embeddings extracted from an image is not only disentangled, but also interpretable and coherent.

\subsection{artistic image generative learning stage}
\label{sec:inject}
In the generative learning stage, We aim to feed the content and style embeddings of CSDN to a frozen SD model for controllable image generation. First, we project disentangled style and content representations $e_s$ and $e_c$ into the feature dimensions required by the SD model using a projection layer, resulting in the style embeddings and content embeddings for SD, $e^{SD}_s$ and $e^{SD}_c$, respectively. Then, we use multi-step learnable cross-attention layers (MCL) to inject the style and content embeddings into the denoising process of the SD model. At each timestep of the diffusion process, the style and content embeddings are introduced as conditions through the cross-attention layers in MCL to guide the generation process. These cross-attention layers embed the style and content embeddings into the current diffusion features using the attention mechanism:
\begin{align}
    &Q = W_{Q}Z \\
    &K =
    \begin{cases} 
        e^{SD}_cW_{c}^{K}, & \text{} \\ 
        Concat(e^{SD}_cW_{c}^{K}, e^{SD}_sW_{s}^{K}), & \text{middle block}
    \end{cases} \\
    &V =
    \begin{cases} 
        e^{SD}_cW_{c}^{V}, & \text{} \\ 
        Concat(e^{SD}_cW_{c}^{V}, e^{SD}_sW_{s}^{V}), & \text{middle block}
    \end{cases} \\
    &Z_{new} = Softmax(\frac{QK^T}{\sqrt{d}})V
\end{align}
where $Z$ and $Z_{new}$ represent the noise states at the current step and the next step during the denoising process, respectively. Inspired by \cite{wang2024instantstyle}, we inject the style embeddings only into the middle block of U-Net, which also benefits for preventing content leak.
For the content text, we first extract text features using the original text encoder from SD, then concatenate the text features with the disentangled content representations $e_c$ extracted by CSDN, before injecting them into the diffusion process.

SD model initially transforms an input image  $\mathit{x}$  into a latent code  $\mathit{z}$. The noised latent code  $\mathit{z}_{\mathit{t}}$ at timestep $\mathit{t}$  serves as the input for the denoising U-Net  $\epsilon_{\theta}$ , which interacts with content prompts $\mathit{c}$ and style prompts $\mathit{s}$ through cross-attention. The supervision for this process is ensured by:
\begin{equation}
\mathcal{L}_{\mathit{rec}} = \mathbb{E}_{\mathit{z}, \mathit{c}, \mathit{s}, \epsilon \sim \mathcal{N}(0,1), t} \left[ \left\| \epsilon - \epsilon_{\theta}(\mathit{z}_{\mathit{t}}, \mathit{t}, \mathit{c}, \mathit{s}) \right\|_2^2 \right]
\end{equation}
where $\epsilon \sim \mathcal{N}(\textbf{0},\textbf{I})$ is a noise. The objective is corresponding to the reconstruction loss as in Fig. \ref{fig:model}.

\textbf{Remark:} To prevent the model from becoming complacent during training by simply copying content from image features, which could diminish its generalization ability, we randomly replace the style or content embeddings extracted from images with the corresponding embeddings extracted from texts.
Both content and style embeddings are randomly selected from multiple modalities to participate in training the model. 
The logic behind this approach lies in feeding more complex tasks to train the model is beneficial for improving model's capabilities.

This approach fosters a more nuanced and versatile understanding of both content and style, enabling the model to generate images that are both faithful to the input content and creatively infused with the desired style. Additionally, it also enables our model to generate outputs with various multimodal combinations. Finally, we randomly drop the keywords of style texts for enabling the model to accept a wider variety of style keyword combinations when using text as style prompts.

\section{Experiment}

\subsection{Experiment Settings}

\textbf{Implementation Details.} 
We trained on L20-40G GPUs with 3750 total batches, using AdamW \cite{loshchilov2017decoupled} as the optimizer, with a learning rate of 5e-5, and performed 100 iterations. Regarding reasoning, for guidance without classifiers ~\citep{ho2022classifier}, we use a scale of 7.5 and set T = 50 steps for DDIM ~\citep{song2020denoising} sampling. All comparison methods were implemented using publicly available code and default settings. Detailed settings of our model are elaborated in Supplementary Material Sec.2.

\textbf{Evaluation Metrics} 

Following previous works ~\citep{qi2024deadiff, jiang2024artist}, we evaluate multi-modal stylization using the following metrics:

Content Similarity (TA): Measured by the cosine similarity between stylized images and their corresponding text prompts in the CLIP embedding space.
Image Quality (IQ): Assessed using the LAION-Aesthetics Predictor V2 \href{https://github.com/christophschuhmann/improved-aesthetic-predictor}{1}.
Style Similarity (SS): Defined by a text prompt template "the painter is [v], the theme is [v]", with similarity computed between the generated image and the style prompt in the CLIP text-image embedding space.
Subjective Preference (SP): Evaluated through a user study.


\begin{figure*}[htbp]
    \centering
    \includegraphics[width = \textwidth]{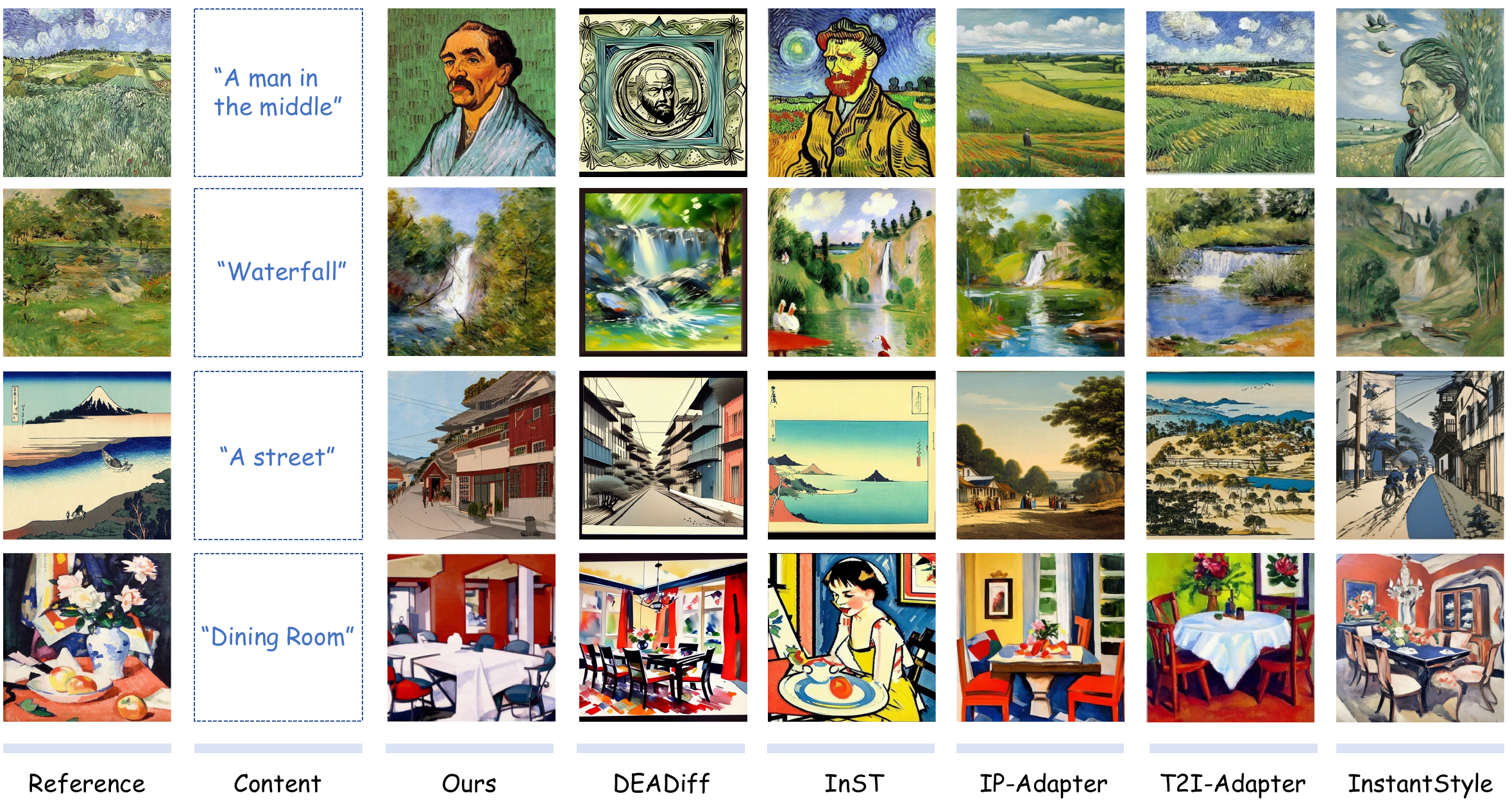}
    \caption{Qualitative comparison with the state-of-the-art text-to-image stylization methods.}
    \label{comparison1}
\end{figure*}

\subsection{Comparison with State-of-the-Arts}

In this section, we compare our method with the state-of-the-art methods. We introduce the experimental results based on supported input modalities and tasks as follows: 1) Text-to-image stylization, including optimization-based methods: InST~\citep{zhang2023inversion}, encoder-based methods: IP-Adapter~\citep{ye2023ip}, DeaDiff~\citep{qi2024deadiff}, T2I-Adapter~\citep{mou2024t2i}, and InstantStyle~\citep{wang2024instantstyle}; 2) Text-to-image generation, featuring advanced models DallE~\citep{ramesh2021zeroshottexttoimagegeneration} and Stable Diffusion (SD)~\citep{rombach2022high}; 3) Collection-based stylization: Artbank~\citep{zhang2024artbank}; 4) Content-style disentanglement: DEADiff~\citep{qi2024deadiff}, StyleDrop \cite{sohn2023styledrop} and DreamStyler \cite{ahn2024dreamstyler}
Due to space limitations, the comparisons for 2-4 are provided in Supplementary Material Sec.5.1-5.3. 


\begin{table*}
  \caption{
  Quantitative comparison with the state-of-the-art text-to-image stylization methods.}
  \vspace{-2mm}
\begin{center}
  \begin{tabular}{c|cccccc}
    \toprule
    {\multirow{1}*{Metrics}}
& InST  & IP-Adapter & DEADiff & T2I-Adapter & InstantStyle & Ours\\
    \midrule 
    {SS $\uparrow$} & 0.283 & \underline{0.288} & 0.236 & 0.276 & 0.280 & {\textbf{0.293}} \\
    {IQ $\uparrow$} & 5.845 & 5.856 & \underline{5.891} & {\textbf{5.895}} & 5.798 & 5.811 \\
    {TA $\uparrow$} & 0.294 & 0.225 & 0.314 & 0.313 & \underline{0.316} & \textbf{0.317} \\
    {SP $\uparrow$} & 2.333 & 2.167 & 2.583 & 2.250 & \underline{3.333} & {\textbf{3.857}} \\
    \bottomrule
  \end{tabular} 
  \end{center}
  \label{table1}

  \vspace{-4mm}
\end{table*}

\subsubsection{Text-to-image stylization}

Fig.\ref{comparison1} illustrates the comparison results with state-of-the-art methods. For methods lacking effective decoupling mechanisms, such as IP-Adapter, T2I-Adapter and InST, semantic conflicts from the reference images are evident in the generated results, as shown in the first and third rows of Fig. \ref{comparison1}. DEADiff and InstantStyle struggle to accurately capture the style of the reference image when there is significant semantic gap between the reference image and the textual prompt. In contrast, our method not only faithfully reflects textual prompts but also aligns with the artistic characteristics of the reference style. For more comparison results, please refer to Supplementary Material Sec 5.4.

\begin{figure*}[t]
    \centering
    \includegraphics[width = 0.9\textwidth]{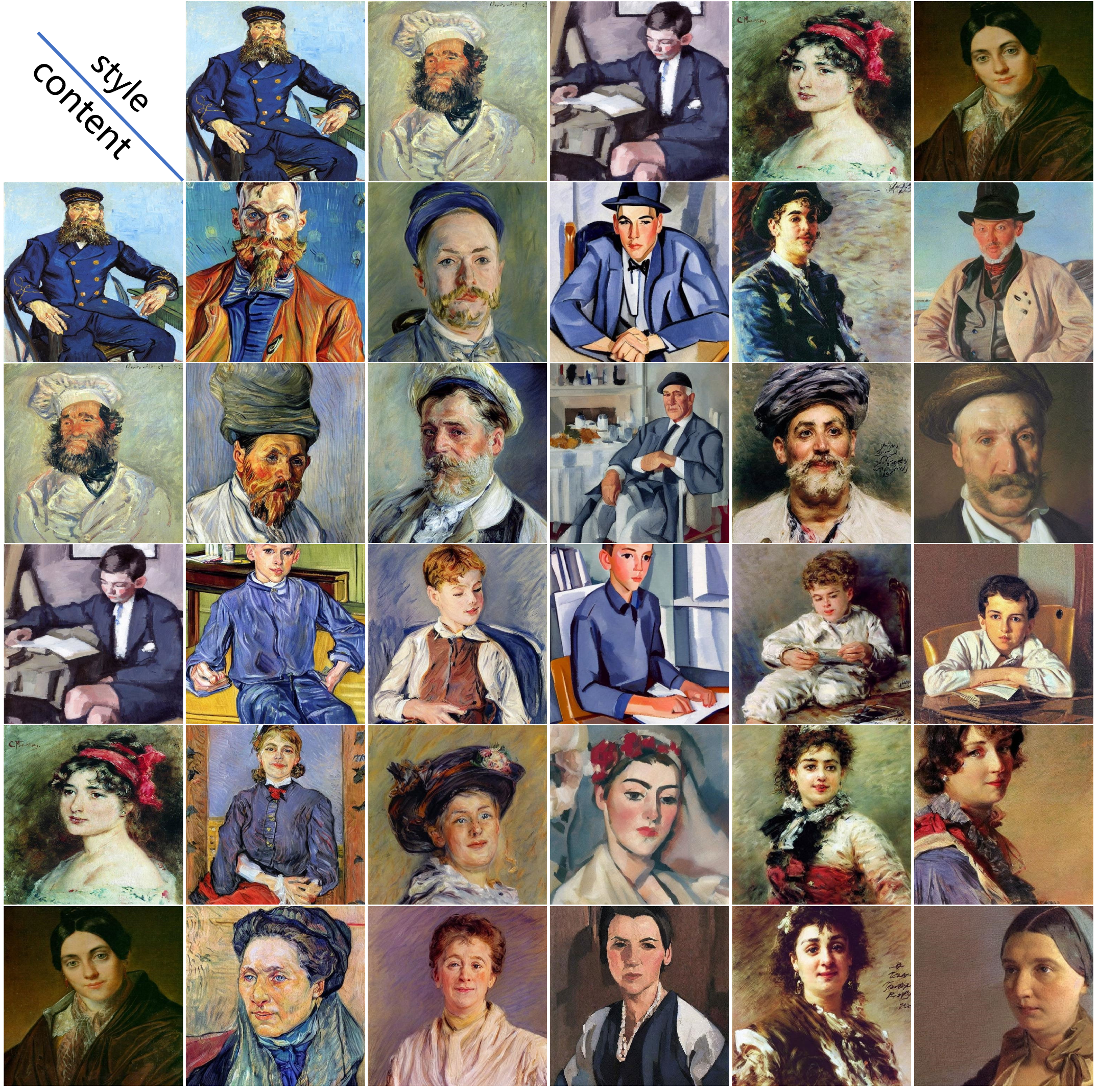}
    \caption{Qualitative results for content and style disentanglement.}
    \label{decouple}
    \vspace{-4mm}
\end{figure*}

\subsubsection{Quantitative Comparison}

Tab. \ref{table1} presents the style similarity, image quality, text alignment and the overall subjective preference of our method compared with the state-of-the-art methods. We can see that our method achieves the highest style similarity and content alignment, demonstrating that our method, through decoupling representations, effectively fixed the problem of semantic conflict and captured the overall artistic style. Furthermore, users demonstrate a significantly greater preference for our method over other ones. More quantitative results are provided in Supplementary Material Sec. 5.5.

\subsection{Content Style Disentanglement} 

\subsubsection{Qualitative illustration of disentanglement effect}

To verify the disentanglement effect, we conducted an experiment using portrait paintings as content and style respectively. As shown in Fig. \ref{decouple}, the generated images depict the content themes accurately: middle-aged men (first two rows), a young boy (third row), and middle-aged women (fourth and fifth rows). Vertically, the styles of the reference images are also well-preserved, with distinct brushstrokes in the third and fourth columns. This demonstrates the robustness of the proposed content-style disentanglement method. We also provided more disentanglement experiments using still life painting as content and style in Supplementary Material Sec. 7.

\subsubsection{Impact of Content and Style Detail Levels on Disentanglement Performance}

\begin{figure}[htbp]
    \centering
    \begin{subfigure}[b]{\linewidth} 
        \centering
        \includegraphics[width=\linewidth]{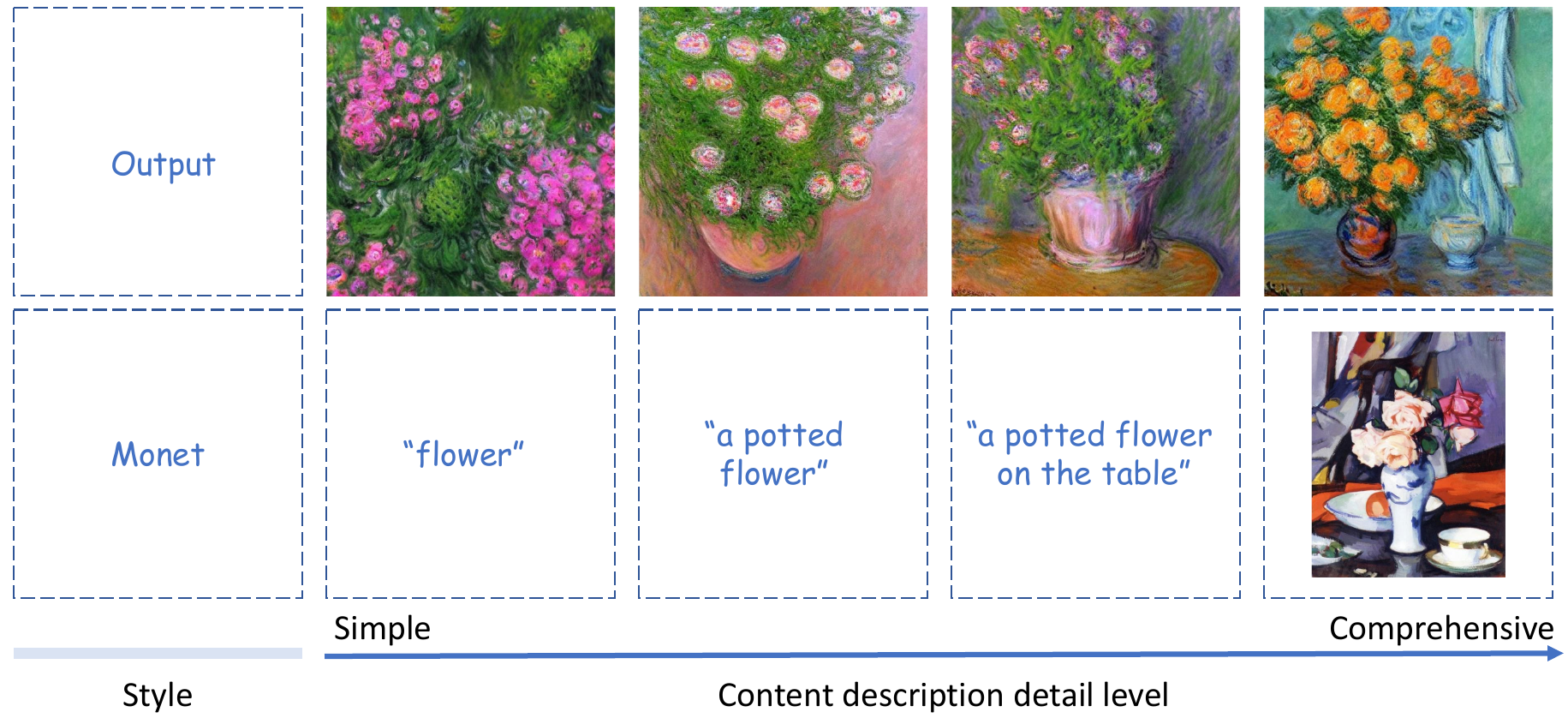} 
        \caption{Impact of content prompt detail levels on disentanglement performance}
        \label{fig:sub1}
    \end{subfigure}
    
    \vspace{0.5cm} 

    \begin{subfigure}[b]{\linewidth} 
        \centering
        \includegraphics[width=\linewidth]{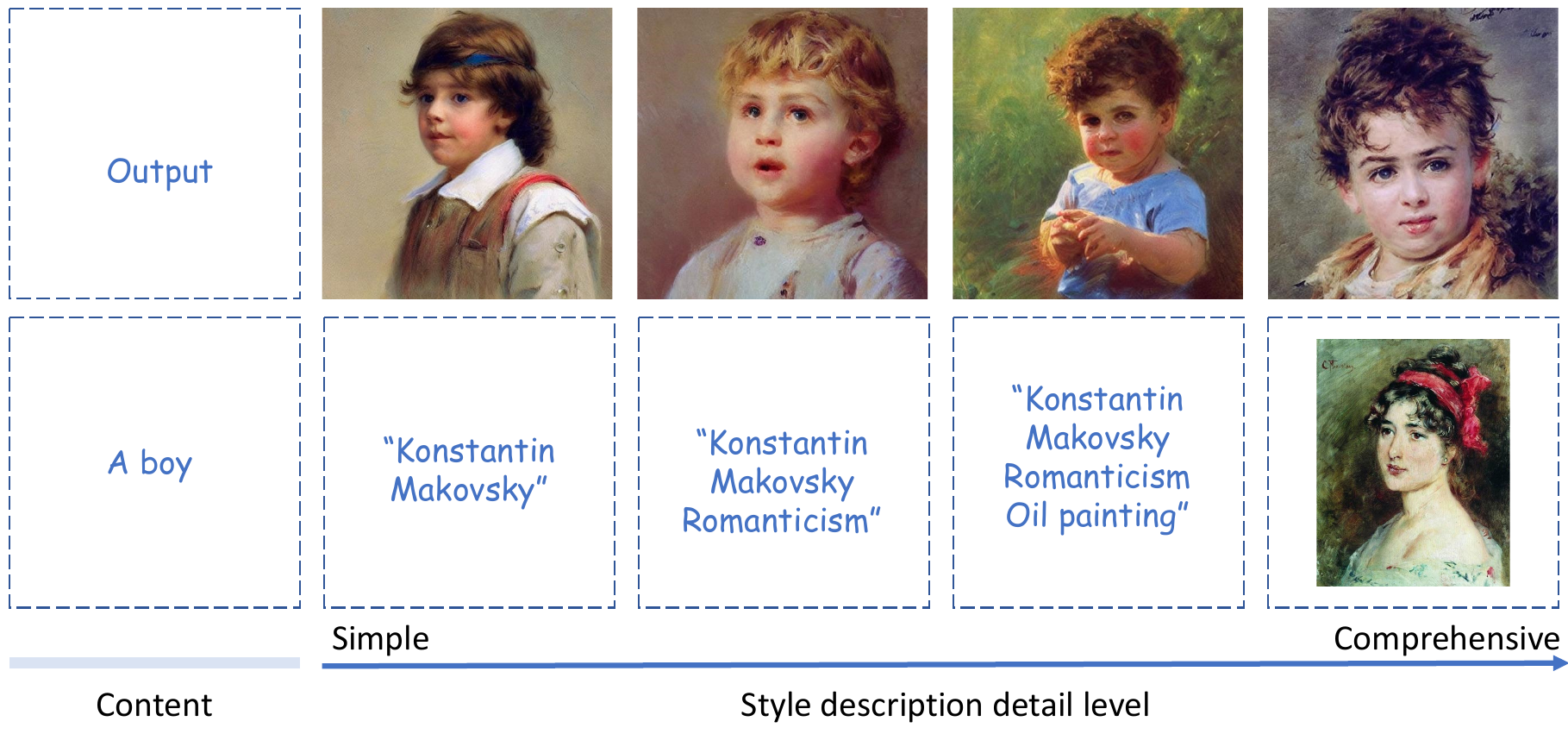} 
        \caption{Impact of style prompt detail levels on disentanglement performance}
        \label{fig:sub2}
    \end{subfigure}

    \caption{Impact of content and style description detail levels on disentanglement performance}
    \label{detail_levels}
\end{figure}

In Figure \ref{detail_levels}, we present the impact of varying levels of detail in content and style descriptions on disentanglement performance. The descriptions for content and style can be provided either through text or images, with information complexity ranging from the simplest to the most intricate—where images provide the highest level of detail.

From subfigure (a), we observe that as textual content descriptions increase in complexity, the generated images accurately reflect the described content. When an artistic image is used for content input, our method effectively disentangles content, capturing key elements like flowers, a vase, a table, and a teacup. Meanwhile, the generated images consistently replicate Monet's brushstrokes and artistic style, regardless of of the level of detail in the content description.


From subfigure (b), we observe that as the textual descriptions of style range from a single word to the artist’s actual paintings, the generated content remains entirely unaffected. However, the level of detail in the generated style increases significantly with more detailed style descriptions.

These results demonstrate the effectiveness of our network design for explicit disentanglement. The robustness of disentanglement is preserved even as the level of detail in content and style descriptions varies, further validating the reliability of our approach.




\subsection{Ablation Study}

\begin{figure}[htbp]
    \centering
    \includegraphics[width = 0.95\linewidth]{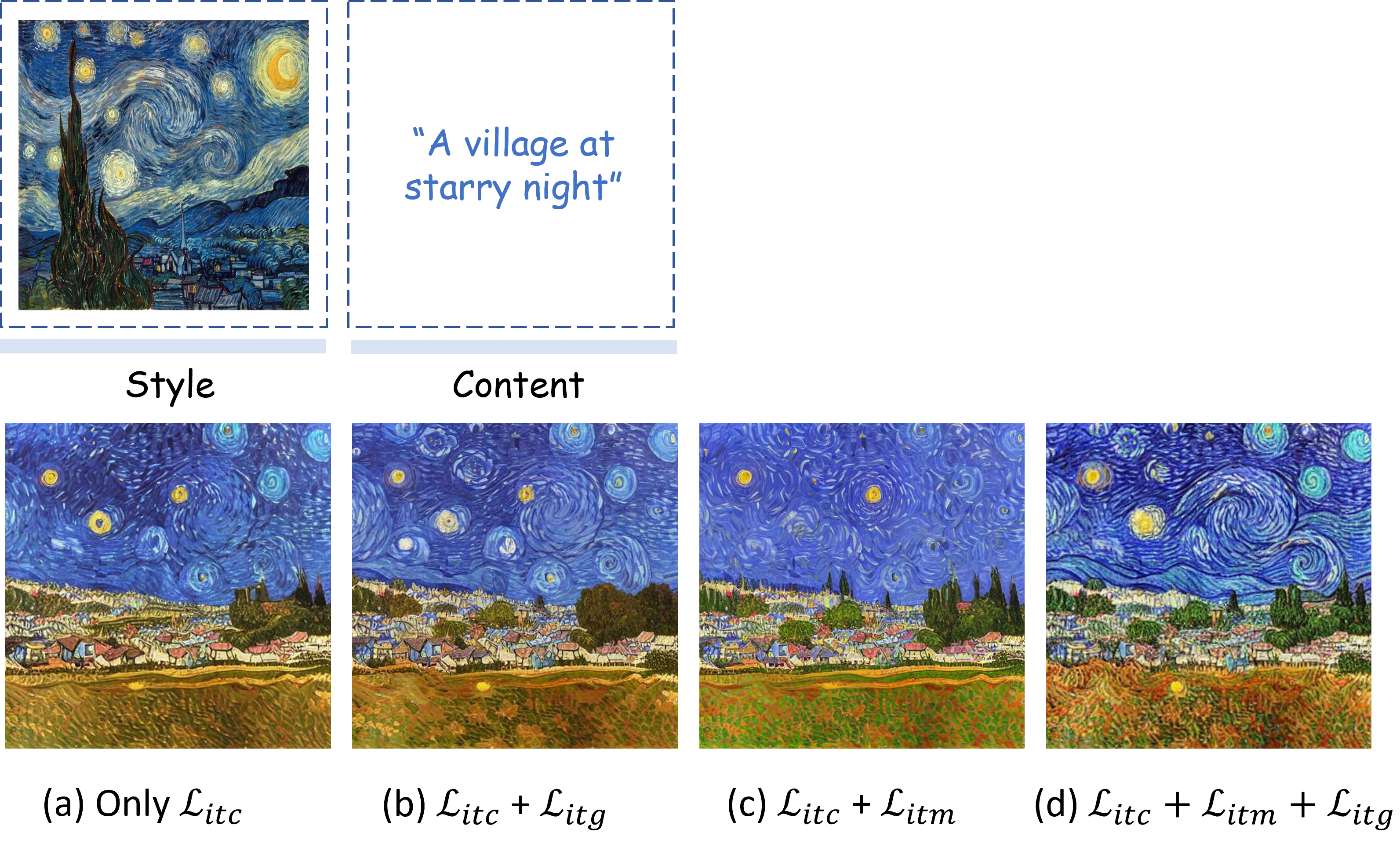}
    \caption{Visual illustration for ablation study on disentanglement loss function components.}
    \label{ablation_visual}
    \vspace{-3mm}
\end{figure}

To evaluate the contribution of each loss function in training the multimodal alignment and disentanglement network, we conducted an ablation study. The results are shown in Figure 
 \ref{ablation_visual}. ITC is the core loss function in the first stage. ITC ensures that the generated images maintain semantic similarity with the text input. Although the styles vary, all images consistently depict a village under a starry night, demonstrating ITC's role in content control. ITG ensures that the model accurately extracts content information from the input image and utilizes it for text-guided generation, enabling more precise stylization. In Figure \ref{ablation_visual} (b), the generated images effectively incorporate Van Gogh's brushstroke style in architectural structures, terrain, and the sky. ITM enhances the alignment between text and images (Figure \ref{ablation_visual} (c)), but at the cost of reduced stylization. Ultimately, these three loss functions work together to ensure that the model accurately represents the content described in the text while preserving the stylization effect, thereby improving text-guided artistic image generation (Figure \ref{ablation_visual} (d)). More ilustration for ablation study is provided in Supplementary Material Sec. 4.


\section{Conclusion}

In this paper, we proposed a multimodal approach for content and style representation disentanglement for artistic image stylization. We constructed a multimodal art dataset, \textit{WikiStyle+}, to provide explicit supervision for decoupling. We employed contrastive learning tasks to learn disentangled content and style representations, which then guided a diffusion model to generate stylized images. Our experiments across various tasks demonstrated the superiority of our method and highlighted the importance of effective content and style disentanglement in image stylization.


\bibliographystyle{unsrtnat}
\bibliography{references}  






\end{document}